\documentclass[twocolumn]{article}
\usepackage{times}
\usepackage{graphicx}

\usepackage{array}

\usepackage{url}
\usepackage{paralist}
\usepackage[small]{caption}

\newcolumntype{M}[1]{>{\centering\arraybackslash}m{#1}}
\newcolumntype{N}{@{}m{0pt}@{}}

\title{Responsible Autonomy}
\author{Virginia Dignum\\ Delft University of Technology \\ m.v.dignum@tudelft.nl}

\begin{document}

\maketitle

\begin{abstract}
As intelligent systems are increasingly making decisions that directly affect society, perhaps the most important upcoming research direction in AI is to rethink the ethical implications of their actions. Means are needed to integrate moral, societal and legal values with technological developments in AI, both during the design process as well as part of the deliberation algorithms employed by these systems. In this paper, we describe leading ethics theories and propose alternative ways to ensure ethical behavior by artificial systems. 
Given that ethics are dependent on the socio-cultural context and are often only implicit in deliberation processes, methodologies are needed to elicit the values held by designers and stakeholders, and to make these explicit leading to better understanding and trust on artificial autonomous systems.
\end{abstract}

\section{Introduction}\label{sec:intro}
It is no news that Artificial Intelligence (AI) is increasingly entering the public domain in fields such as  transportation, service robots, health-care, education, public safety and security, employment and workplace, and entertainment. Developments in autonomy and learning technologies are rapidly enabling AI systems to decide and act without direct human control. 
As these advances continue at high speed, there is a growing awareness that a responsible approach to AI is needed to ensure the safe, beneficial and fair use AI technologies, to consider the implications of morally relevant decision making by machines, and the ethical and legal consequences and status of AI. 

Design methods and tools to elicit and represent human values, translate these values into technical requirements, and deal with moral overload when numerous values are to be incorporated, are needed to demonstrate that design solutions realize the values wished for. 



As an example, recently much attention has been given to the ethical dilemmas self-driving cars face when needing to deal with potentially life-threatening decisions~\cite{Bonnefon1573}. This has been described as an application of the well-known \textit{trolley problem}~\cite{foot1967problem}, an hypothetical scenario, long used in philosophy and ethics discussions, where a runway trolley is speeding down a track to which five people are tied up and unable to move. An observer has control over a lever that can switch tracks before the trolley hits the five people, but also on this alternative track there is one person tied to the tracks. The moral dilemma concerns the decision the observer should make: do nothing and allow the trolley to kill the five, or pull the lever and kill one? This is obviously a hypothetical scenario without any direct practical application. However, it can be seen as an abstraction for many dilemmas involving AI systems now and in the future. Besides the relation to self-driving cars, similar dilemmas will need to be solved by intelligent medicine dispenses faced with the need to choose between two patients when it does not have enough of a needed medicine, by search and rescue robots faced with the need to prioritize victims, or, as we have recently shown, by health-care robots needing to choose between user's desires and optimal care~\cite{pizza2017icai}. In this paper, we use the trolley scenario as illustration of this wide application of moral dilemmas deliberation.\\
From an ethical perspective, there are no optimal solutions to these dilemmas. In fact, different ethical theories will lead to distinct solutions given that values hold by individuals, groups and societies put different preferences on the action to choose. Understanding of these differences is essential to the design of AI systems able to deal with such dilemmas. Moreover, it is not so that the responsibility lays solely with the individual (human or machine) that operates the lever: societal, legal and physical infrastructures are also means to determine the decision.

If we are to build AI systems that can deal with this type of ethical dilemma, and ensure that AI is developed responsibly incorporating social and ethical values, societal concerns about the ethics of AI must be reflected in design.  AI systems should therefore be ground on  principles of Accountability, Responsibility and Transparency (ART), extending and characterizing the classic  principles of Autonomy, Interactivity and Adaptability described in \cite{floridi2004morality,russell2009norvig}.

Firstly, \textit{Accountability} for the decision must be derivable from the algorithms and data used by the system in order to make the decision. This includes the need for representation of the moral values and societal norms holding in the context of operation, which the agent uses for deliberation. Secondly, even if the AI system is the direct cause of action, the chain of \textit{Responsibility} must be clear, linking the agent's decision to user, owner, manufacturer, developer, and all other stakeholders whose actions in one way or another contribute to the decision. 
Finally, explanation of actions requires \textit{Transparency} in terms of the algorithms and data used, their provenance and their dynamics. I.e., algorithms must be designed in ways that let us inspect their workings.

This paper is organized as follows. Section \ref{sec:rai} positions this work within the topic of Responsible Artificial Intelligence, discussing in particular the Value-Sensitive Design methodology and  the development of Artificial Moral Agents.  We describe relevant ethical theories in Section \ref{sec:ethics}. In Section \ref{sec:design} we discuss how ethical reasoning in AI can differ based on these theories, and the role of value systems in moral deliberation. 
In Section \ref{sec:reasoning}, we discuss how the integration of Ethical theories and Value systems lead to different responses to moral dilemmas, and propose mechanisms for implementation. Finally, in Section \ref{sec:conclusions} we present preliminary conclusions and directions for further research.



\section{Responsible Artificial Intelligence}\label{sec:rai}

Central premise of Responsible AI is that in order for AI systems to be safe, accepted and trusted, the system should be designed to take ethical considerations into account and to consider the moral consequences of its actions and decisions, in accountable, responsible, and transparent ways. Only then, their goals, their decisions, and the actions they take to achieve these, will be closely aligned with human values. 

Ethical considerations in the development of intelligent interactive systems is becoming one of the main influential areas of research in AI, and has led to several initiatives both from researchers as from practitioners, including the IEEE initiative on Ethics of Autonomous Systems\footnote{\url{http://standards.ieee.org/develop/indconn/ec/autonomous_systems.html}}, the Foundation for Responsible Robotics\footnote{\url{http://responsiblerobotics.org/}}, 
and the Partnership on AI\footnote{\url{http://www.partnershiponai.org/}} which brings together the largest tech companies to advance public understanding and awareness of AI and its potential benefits and costs.

\subsection{Design for Values}\label{sec:vsd}
Responsible AI starts with design processes that ensure that design decisions are formulated explicitly rather than being implicit in the procedures and objects. In particular, the the values and value priorities of designers and stakeholders should be elicited in a participatory way that ensures that global aims and policies are clear, shared, and context-oriented.

Value-Sensitive Design (VSD) methodologies, also known as Design for Values or Values in Design, are technology design approaches that take human values as the central focus of design \cite{friedman2006,Hoven05}. As such, VSD is an ideal candidate for the design of AI technology. The underlaying premise is design is never value free, and the identification, and explicit representation of the values underlying design leads to better designs. Value-sensitive design enables engineers and developers to give conflicting social values a place in smart design and to combine them is such a way as to reach a win-win situation. 

Value sensitive design is a theoretically grounded approach to the design of technology that accounts for human values in a principled, systematic and comprehensive manner. It situates moral questions early on in the process of design, development of technologies, systems and research. It proposes rational procedures for designing artifacts under the guidance of moral values. 
VSD frameworks bridge the gap between abstract values and concrete system implementation. VSD aims at the translation of values into tangible design requirements. Of particular relevance for the design of AI systems are value hierarchies, i.e. a hierarchical structure of values, general norms and more specific design requirements or goals~\cite{vandePoel2013}. A values hierarchy provides a \textit{for-the-sake-of} specification link, describing how values are translated into norms, into requirements making explicit the design decision. In the opposite direction, links form an explicit constitutive relation \cite{searle2010making}, indicating which goal \textit{counts-as} a norm, counts-as a value in a given context. 


\subsection{Levels of Morality}\label{sec:levels}

\begin{table*}[t]
\centering
\caption{Comparison of Main Ethical Theories}\label{table:ethics}
\begin{tabular}{|M{1.8cm}|M{4.8cm}|M{4.8cm}|M{4.8cm}|N} \hline
&\textbf{Consequentialism}&\textbf{Deontology}&\textbf{Virtue Ethics}&\\[10pt] \hline
\textbf{Description}& An action is right if it promotes the best consequences, i.e where happiness is maximized.& An action is right if it is in accordance with a moral rule or principle. 
 & An action is right if it is what a virtuous agent would do in the circumstances.  &\\[10pt] \hline
\textbf{Central \newline Issue} & The results matter, not the actions themselves & Persons must be ends in and of themselves and may never be used as means & Emphasize the character of the agent making the actions &\\[10pt] \hline
\textbf{Guiding Value}& Good (often seen as maximum happiness) & Right (rationality is doing one's moral duty) & Virtue (dispositions leading to the attainment of happiness) &\\[10pt] \hline
\textbf{Practical \newline Reasoning} & The best for most (means-ends reasoning)& Follow the rule 
(rational reasoning) & Practice human qualities (social practice)&\\[10pt] \hline
\textbf{Deliberation Focus} & Consequences (What is outcome of action?) & Action (Is action compatible with imperative?) & Motives (is action motivated by virtue?) &\\[10pt] \hline
\end{tabular}
\end{table*}

Even though, being pieces of software and hardware, AI systems are basically tools, their increased intelligence and inter-ability capabilities makes that AI systems are increasingly being perceived, and expected to behave, as partners to their users, with the duties and responsibilities we expect from human teammates \cite{casa12}. 
\cite{wallach2008moral} proposes a pathway to engineering ethics in AI comprising \textit{operational, functional and full ethical} behaviour.  At the lowest level of ethical behaviour, \textbf{Tools}, such as search engines, do not have either autonomy nor social awareness and are not considered to be ethical systems, but incorporate in their design the values of their engineers, and are therefore said to have \textit{operational morality}. As system autonomy and social awareness increases, \textbf{Assistant} systems, are able to act independently in open environments with \textit{functional morality}, i.e. are sensitive to ethically relevant features of their environment, based on hard-wired ethical rules, resulting in autonomous agents that are able to adjust their actions to human norms. Most normative systems fall into this category \cite{conte1998autonomous,boella2006introduction}. Finally, \textbf{Artificial Moral Agents (AMA)} are able of self-reflection and can reason, argue and adjust their moral behavior to that of their partners and context. 

\section{Ethics for AI}\label{sec:ethics}
In order to build machines that follow ethical principles, we first need to understand the different ethical theories that can be applied to decision-making. Note that this paper  focuses on ethical deliberation by AI systems, and not on other areas of AI Ethics such as regulation and codes of conduct, and AI and robot rights.



Ethics (or Moral Philosophy) is concerned with questions of how people ought to act, and the search for a definition of right conduct (identified as the one causing the greatest good) and the good life (in the sense of a life worth living or a life that is satisfying or happy). From the perspective of understanding and applying ethical principles to the design of artificial systems, Normative Ethics (or Prescriptive Ethics) are of particular relevance. Normative ethics is the branch of ethics concerned with establishing how things should or ought to be, how to value them, which things are good or bad, and which actions are right or wrong. It attempts to develop a set of rules governing human conduct, or a set of norms for action. 
In the following, we briefly  introduce Consequentialism, Deontology and Virtue Ethics as exemplary of the main schools of thought in Normative Ethics. For more information on Normative Ethics, we refer to e.g. the Stanford Encyclopedia of Philosophy\footnote{\url{https://plato.stanford.edu}}.
The aim being to show their different impact on possible agent deliberation approaches and to be extensive. %
Normative ethical theories can be categorized into three main categories: .  

\textbf{Consequentialism} (or Teleological Ethics) argues that the morality of an action is contingent on the action's outcome or result. Thus, a morally right action is one that produces a good outcome or consequence. Consequentialist theories must consider questions like ``What sort of consequences count as good consequences?", ``Who is the primary beneficiary of moral action?", ``How are the consequences judged and who judges them?"

\textbf{Deontology}  is the normative ethical position that judges the morality of an action based on rules. This approach to ethics focuses on the rightness or wrongness of the action description that is used in the decision to act, as opposed to the rightness or wrongness of the consequences of those actions. It argues that decisions should be made considering the factors of one's duties and other's rights. Deontologic systems are about having a set of rules to follow, i.e. can be seen as a top-down approach to morality. Kant's Categorical Imperative roots morality in the rational capacities of people and asserts certain inviolable moral laws. Kant argues that to act in the morally right way, people must act according to duty, and that it is the motives of the person who carries out the action that make them right or wrong, not the consequences of the actions.


Finally, \textbf{Virtue Ethics}, focuses on the inherent character of a person rather than on the nature or consequences of specific actions performed. This theory identifies virtues (those habits and behaviours that will allow a person to achieve well being or a good life), counsels practical wisdom to resolve any conflicts between virtues, and claims that a lifetime of practicing these virtues leads to, or in effect constitutes, happiness and the good life. Virtue ethics indicate that \textit{regret} is an appropriate response to a moral dilemma. \\

Note that our aim is not to provide a full landscape of Ethical theories but to present the most exemplary alternatives applicable in AI reasoning. Other approaches suitable for AI, such as the \textit{principle of double effect} (DDE), the  \textit{principle of lesser evils} and  \textit{human rights ethics}, can be seen as alternatives to the exemplary theories described above. 

These and other ethical theories are currently being considered on the discussion around the governance and legal position of AI 
and have led to concrete proposals such as that of the Engineering and Physical Sciences Research Council (EPSRC)  in the UK, listing a set of principles for designers, builders and users of robots in the real world\footnote{See: \url{https://www.epsrc.ac.uk/research/ourportfolio/themes/engineering/activities/principlesofrobotics/}}, or the one currently under discussion by the European Parliament. 

Table \ref{table:ethics} gives a comparison of these Ethics theories. AI systems that can deal with ethical reasoning should meet the following requirements, further discussed in Section~\ref{sec:reasoning}:
\begin{itemize}
\item Representation languages rich enough to link domain knowledge and agent actions to the `Value' central to the theory; 
\item Planning mechanisms appropriate to the Practical Reasoning prescribed by the theory
\item Deliberation capabilities to deal with the Focus of the theory.
\end{itemize}
Obviously, many architectures are possible that meet these requirements, and more research is needed to further elaborate on these issues. Our aim here is to provide a sketch of the possibilities rather than a full account of architectural and implementation characteristics. In  Section \ref{sec:reasoning}, we describe the effect of the different Ethics theories on the results of deliberation, taking as scenario the trolley problem introduced in Section \ref{sec:intro}.

\section{Design for Responsible Autonomy}\label{sec:design}
Ethical theories provide an abstract account of the motives, questions and aims of moral reasoning. For its practical application, more is needed, namely how and by who deliberation is done, and understanding which moral and societal values are at the basis of deliberation. E.g. Consequentialistic approaches aim at `the best for the most' but one needs to understand societal values in other to determine what counts as the `best'. In fact, depending on the situation, this can be wealth, health, sustainability or another value. In this section, we turn our attention to the design of AI systems.
We present several design options concerning who is responsible to the decision, and how decisions are dependent on the relative priority of different moral and societal values.


\subsection{Who takes the decision?}\label{sec:who}
Even though most work in Artificial Moral Agents (AMA) refers to automated decision-making by the machine itself, in reality the spectrum of decision making is much wider, and in many cases the actual decision by the machine itself is limited. Depending on the level of autonomy and regulation, we identify four possible approaches to design decision-making mechanisms for autonomous systems and indicate how these can be used for moral reasoning by the AI systems described in Section \ref{sec:levels}: 

\begin{itemize}
\item \textbf{Human control:} in this case a person or group of persons are responsible for the decision. Different control levels can be identified from that of a auto-pilot, where the system is in control and the human supervises, to that of a `guardian angel' where the system supervises human action. 
From a design perspective, this approach requires to include means to ensure shared awareness of the situation, such that the person taking decision has enough information at the time she must intervene. Such interactive control systems are also known as human-in-the-loop control systems~\cite{Li2014}. This is the decision-making mechanism required for Tools.
\item \textbf{Regulation:} here the decision is incorporated, or constrained in the systemic infrastructure of the environment. In this case, the environment ensures that the system never gets into moral dilemma situation. 
I.e the environment is regulated in such ways that deviation is made impossible, and therefore moral decisions by the autonomous system are not needed. This is the mechanism used in smart highways, linking road vehicles to their physical surroundings, where the road infrastructure controls the vehicles \cite{misener2006path}. In this case, ethics are modeled as regulations and constraints to enable that systems can suffice with limited moral reasoning, as is the case of Assistants in the categorization in section \ref{sec:levels}.
\item \textbf{Artificial Moral Agents (AMA):} these are AI systems able to incorporate moral reasoning in their deliberation and to explain their behaviour in terms of moral concepts. An AMA \cite{wallach2008moral} can autonomously evaluate the moral and societal consequences of its decisions and use this evaluation in their decision-making process. Here moral refers to principles regarding \textit{right} and \textit{wrong}, 
and explanation refers to algorithmic mechanisms to provide a qualitative understanding of the relationship between the system's beliefs and its decisions. This approach requires complex decision making algorithms, based e.g. on deontic logics, and/or reinforcement learning. These mechanisms ensure Full ethical behavior required by Partner systems.
\item \textbf{Random:} the autonomous system randomly chooses its course of action when faced with a (moral) decision. The claim here is that if it is ethically problematic to choose between two wrongs, an elegant solution is to simply not make a deliberate choice \footnote{cf. Wired: \url{https://goo.gl/FGKhE5}}. The Random mechanism can be seen as an approximation to human behavior and can be applied to any type of system. Interestingly, there is some empirical evidence that, under time pressure, people tend to choose for justice and fairness over careful reasoning \cite{SJOP:SJOP367}. This behaviour could be implemented as a weak form of randomness. 
\end{itemize}

These four classes of decision-makers differ in terms of Accountability, Responsibility and Transparency (ART), where Accountability is related to answerability, blameworthiness and liability, Responsibility refers to being in charge, or being the cause behind whether something succeeds or fails, and Transparency includes openness of data, processes and results for inspection and monitoring. In most situations, accountability and responsibility come together, i.e. not possible to have one without the other. According to these definitions, it can be claimed that the algorithmic approach above can be described as not accountable but responsible, and that for the Random approach nor Accountability nor Responsibility are possible. I.e. even if the (machine-learning) algorithms behind the decision cannot explain (or answer for) the decision, one can point to the machine as the cause of the decision, and therefore claim it to be the directly responsible entity. In this case, Transparency is of utmost importance to enable trust. This view on AI is the opposite situation of the classic case from business world, where CEO's have claimed to be accountable, i.e. answerable for business decisions but not responsible for possible fraud \cite{Lay11}. Although outside the scope of this paper, we are currently working on  formal semantics representation of the ART concepts as means to explore the issue of legal personhood of autonomous systems, as e.g. currently under discussion in the European Parliament\footnote{cf. \url{http://www.europarl.europa.eu/news/en/news-room/20170110IPR57613}}.



\subsection{Who sets the values?}\label{sec:which}
One of the main challenges for moral reasoning is to determine which moral values to aim for and which ethical principles to adhere to in a given circumstance. Each individual and socio-cultural environment prioritizes different moral and societal values. Therefore, besides understanding how moral decisions are taken, using Ethical theories, another aspect to consider are the cultural and individual values of the people and societies involved. Schwartz has demonstrated that moral values are quite consistent across cultures \cite{Schwartz2006} but that cultures prioritize these values differently \cite{schwartz2006culture,hofstede2001culture}. Basic values refer to desirable goals that motivate action and transcend specific actions and situations, and can be classified along four dimensions:
\begin{inparaenum}[(i)]
\item Openness to change, 
\item Self-enhancement, 
\item Conservation,
\item Self-transcendence.
\end{inparaenum}
As such, values serve as criteria to guide the selection or evaluation of actions, taking into account the relative priority of values. 

Different value priorities will lead to different decisions in a self-driving vehicle dilemma scenario. E.g. a preference for Hedonism will more likely lead to a choose actions that protect the passenger, while Benevolence can lead to prefer actions that protect the pedestrians. 

It is therefore important to identify holding societal values, when determining the rules for moral deliberation by AMAs. Approaches based on crowd-sourcing or direct democracy, can be used to elicit the values of the community, but should be taken with caution. In fact, as in the emperor interpreting the crowd's bidding at the circus, \textit{social acceptance} does not always imply \textit{moral acceptability} and vice-versa \cite{verdiesen2016mood}. In \cite{malle2015sacrifice} and \cite{Bonnefon1573} a social acceptance approach was followed to determine the most appropriate action by a robot in the trolley problem. This has identified that people take different choices when put in the place of the public or that of the vehicle owner. Moral acceptability can be determined using e.g. the Moral Foundations Questionnaire \cite{graham2011mapping} which measures several ethical principles, including \textit{harm}, \textit{fairness} and \textit{authority}.

A combination with morally studies, e.g. according to Ethical systems as those described in section \ref{sec:ethics} and the elicitation of holding community values can be of use here.

\section{Implementing Ethical Deliberation}\label{sec:reasoning}
In this section, we will discuss the engineering of ethical deliberation mechanisms based on the different approaches described in Section \ref{sec:who} and how these meet the ART principles, and provide discuss the effects of implementation of the different Ethical theories presented in Section \ref{sec:ethics} on the deliberation of AMAs.

Assuming that the development of AI systems follows a standard engineering cycle of Analysis - Design - Implement - Evaluate, taking a Design for Values approach basically means that the Analysis phase will need to include activities for 
\begin{inparaenum}[(i)] 
\item the identification of societal values, 
\item deciding on moral deliberation approach (User control, Regulation or AMA), and
\item methods to link values to formal system requirements, such as e.g. \cite{vandePoel2013} or \cite{aldewereld2010making}.
\end{inparaenum}

\begin{table*}[t]
\centering
\caption{Computational and ART consequences of ethical deliberation mechanisms}\label{table:reasoning}
\begin{tabular}{|M{2.3cm}|M{8cm}|M{6.5cm}|N} \hline
\textbf{Ethical Deliberation}&\textbf{Computational reqs}&\textbf{ART}&\\[10pt] \hline
\textbf{User Control} &
\begin{itemize}
\item Realtime reasoning
\item Ensure situational awareness to user
\item Explanation capabilities
\item Output internal state in user understable way
\end{itemize} 
& \begin{itemize}
\item Delegated to user 
\end{itemize} 
&\\[10pt] \hline
\textbf{Regulation} & 
\begin{itemize}
\item Formal link from values to norms to behaviour
\item Define institutions for monitoring and control
\item Moral reasoning can be done off-line
\end{itemize} 
& 
\begin{itemize}
\item A: institutional
\item R: institutional
\item T: system (by requirement)
\end{itemize} 
&\\[10pt] \hline
\textbf{AMA} & \begin{itemize}
\item Formal link from values to norms to behaviour
\item Define reasoning rules
\item Supervised learning of morality
\item Realtime reasoning
\end{itemize} 
& 
\begin{itemize}
\item A: system (by explanation)
\item R: system (by deliberation)
\item T: system (by requirement)
\end{itemize} 
&\\[10pt] \hline

\end{tabular}
\end{table*}
Concerning how moral deliberation mechanisms could be implemented in AI systems it should be noted that moral dilemmas do not have one optimal solution, and the dilemma is exactly how to choose between two `bad' options. As an abstract example of the many morally-oriented decisions that AMAs will need to take in all types of domains and situations, we use the classic trolley problem scenario as introduced in Section \ref{sec:intro}. 
I.e. here, the trolley scenario should be seen as a 
metaphor to highlight many of ethical aspects of choices by machines, such as autonomous vehicles, or care robots. 
The application of ethical theories to the trolley problem leads to different decisions. E.g. taking a Utilitarian, or Consequentialist, approach, the decision would be to save the largest amount of lives, whereas the application of a Human Rights approach would lead to a decision not to switch the lever, as it is not to one to decide on the lives of others, given that each live is valuable in itself \cite{PAPA:PAPA017}. 
Moreover, to design an ethical deliberation mechanism, both Ethics (cf. Section~\ref{sec:ethics}) and Values (cf. Section~\ref{sec:which}) must be considered together. I.e, the agent's capability to evaluate the context and its `personality' will identify different orderings of the values and which ethical theory is most salient in a given situation. Also the order in which different aspects are evaluated or rules are used, can lead to very diverse decisions. These ordering itself is also determined by the values and ethical principles that the system follows, and is influenced by the context of operation. 
For instance when Conservation is the priority value, a decision based on Consequentialism theories would lead to save the largest amount of lives, whereas a Deontological approach would consider traffic laws and possibly also higher level legal systems, such as the human rights, whereas a Virtuous system would choose to take no action (as it would benefit a virtuous person to do no harm deliberately). 

From an implementation perspective, 
the different Ethical theories differ in terms of computational complexity of the required deliberation algorithms. To implement Consequentialist agents, reasoning about the consequences of actions is needed, which can be supported by e.g. dynamic logics. For Deontologic agents, higher order reasoning is needed to reason about the actions themselves. I.e. the agent must be aware of its own action capabilities and their relations to institutional norms, requiring e.g. Deontic logics. Finally, Virtue agents need to reason about its own motives, which lead to actions, which lead to consequences, which are complexer modalities and require constructs to deal with regret and creativity to apply learned solutions to new dilemmas. 

All approaches raise their own specific computational problems, but they also raise a common problem of whether any computer (or human, for that matter) could ever gather and compare all the information that would be necessary for the theories to be applied in real time \cite{allen2005artificial}. This problem seems especially acute for a Consequentialist approach, since the consequences of any action are essentially unbounded in space or time, and therefore a pragmatic decision must be taken on how far should the system go in evaluating possible consequences. The problem does not go away for Deontologic or Virtues approaches because consistency between the duties can typically also only be assessed through their effects in space and time. Reinforcement learning techniques can be applied as means to analyze the evolution and adaptation of ethical behaviour, but this requires further research.

Most importantly is to understand how society will accept these decisions, and how the ART principles differently. In an empirical experiment, Malle has found out ``differences both in the norms people impose on robots (expecting action over inaction) and the blame people assign to robots (less for acting, and more for failing to act)" \cite{malle2015sacrifice}. As illustration, Table \ref{{table:reasoning}} provides a small illustration of the computational issues of the different moral deliberation approaches and how ART can be addressed.  However, further research is needed to understand which are the differences in acceptance of decisions driven by different approaches. 

\textit{Accountability} requires 
both the function of guiding action (by forming beliefs and making decisions), and the function of explanation (by placing decisions in a broader context and by classifying them along moral values). To this effect, machine learning techniques can be used to classify  states or action as `right' or `wrong', basically in the same way as classifiers learn to distinguish between cats and dogs. Another approach to develop explanations methods 
is to apply evolutionary ethics \cite{binmore2005natural} and structured argumentation models \cite{prakken2013}. This enables to create a modular explanation tree where each node explains nodes at lower levels, where each node encapsulate a specific reasoning modules, treated each as a black-box. This moreover provides a model-agnostic approach potentially able to deal with \textit{Transparency} in stochastic, logic and data-based models in a uniform way. Further research is needed to verify this approach. Yet another approach is proposed in ~\cite{gigerenzer2010moral} based on pragmatic social heuristics instead of moral rules or maximization principles. This approach takes a learning perspective integrating both the initial ethical deliberation rules with adaptation to the context.

\section{Conclusions}\label{sec:conclusions}
In all areas of application, AI reasoning must be able to take into account societal values, moral and ethical considerations, weigh the respective priorities of values held by different stakeholders and in multiple multicultural contexts, explain its reasoning, and guarantee transparency. As the capabilities for autonomous decision making grow, perhaps the most important issue to consider is the need to rethink responsibility. 
There is an urgent need to identify and formalize what autonomy and responsibility exactly mean when applied to machines. Whereas taking a moral agent approach 
placing the whole responsibility with the developer as advocated by some researchers \cite{bryson2011just}, or taking a institutional regulatory approach, the fact is that the chain of responsibility is getting longer. Definitions of control and responsibility are needed that are able to deal with a larger distance between human control and system autonomy.

Nevertheless, increasingly, robots and intelligent agents will be taking decisions that can affect our lives and way of living in smaller or larger ways. Being fundamentally artifacts, AI systems are fully under the control and responsibility of their owners  or users. However,  developments in autonomy and learning are rapidly enabling AI systems to decide and act without direct human control. That is,
in dynamic environments, their adaptability capabilities can lead to situations in which the consequences of their decisions and actions will not be always possible to direct or predict.


In this paper, we proposed possible ways to implement ethics and human values into AI design. In particular, we propose several approaches to responsibility: as a task of the human-in-the-loop, as part of the decision-making algorithm, or as part of the the social, legal and physical infrastructures that enable interaction. 


\clearpage \newpage

\bibliographystyle{plain}
\bibliography{refs}

\end{document}